# Comprehensive Study Dynamic CNN

Kamal Sherawat, Vikrant Bhati
Virginia Tech, ECE 6524 - Deep Learning
{kamal}@vt.edu

## Abstract

*This project provides a comparative study of dynamic convolutional neural networks (CNNs) for various tasks, including image classification, segmentation, and time series analysis. Based on the ResNet-18 architecture, we compare five variants of CNNs: the vanilla CNN, the hard attention-based CNN, the soft attention-based CNN with local (pixel-wise) and global (image-wise) feature attention, and the omni-directional CNN (ODConv). Experiments on Tiny ImageNet, Pascal VOC, and the UCR Time Series Classification Archive illustrate that attention mechanisms and dynamic convolution methods consistently exceed conventional CNNs in accuracy, efficiency, and computational performance. ODConv was especially effective on morphologically complex images by being able to dynamically adjust to varying spatial patterns.*

*Dynamic CNNs enhanced feature representation and cross-task generalization through adaptive kernel modulation. This project provides perspectives on advanced CNN design architecture for multiplexed data modalities and indicates promising directions in neural network engineering.*

## 1. Introduction

Convolutional Neural Networks (CNNs) have been revolutionizing the computer vision and pattern recognition area with impressive performance in various applications like image classification, segmentation, and time series analysis. However, for all their popularity, conventional CNNs are constrained by fixed architecture and fixed convolutional kernels, potentially weakening their capacity to cope with the multiplicity and dynamic nature of real-world data. The introduction of more complex models such as AlexNet ,VGG, GoogleNet, ResNet, DenseNet, and Transformer enabled researchers to construct deeper, more accurate, and more efficient models. Most of these well-known deep learning architectures, though, conduct inference statically—the computational architecture and network parameters are fixed after training, restricting their representational capacity, efficiency, and interpretability.

To overcome the above shortcomings, we explore the possibility of extending current models with preserving similar depth and width while making them dynamically change their configurations during the inference phase. We primarily concentrate on Dynamic Convolutional Neural Networks (Dynamic CNNs), which provide a range of benefits in comparison to static architectures:

**Increased Efficiency:** Dynamic CNNs can assign computation adaptively according to the input, cutting down on unnecessary processing for easier samples and concentrating resources on harder cases. This results in drastic improvements in computational efficiency.

**Enhanced Adaptability:** Static CNNs use the same computation for every input, but dynamic CNNs can modify their parameters or architecture in real-time, achieving a trade-off between accuracy and speed according to task requirements.

**More Representation Power:** By dynamically combining multiple convolutional kernels or adapting parameters based on input, dynamic CNNs have the ability to extract complicated patterns and features, which strengthens their learning capability.

**Parameter Efficiency:** Dynamic CNNs achieve more desirable performance–complexity trade-offs, obtaining better accuracy with fewer parameters or computation compared to standard CNNs.

**Enhanced Interpretability:** Dynamic networks can unveil where the model focuses its attention in the input, making decision-making more transparent.

**Compatibility with New Techniques:** Dynamic CNNs can be combined with other deep learning advancements such as attention mechanisms, neural architecture search, and optimization techniques.

In this project we performed a systematic comparative analysis of dynamic convolutional neural networks (CNNs) to investigate these advantages to derive insightful conclusions on the design and application of advanced CNN architectures, showcasing the value of dynamic convolution and attention mechanisms in model flexibility, robustness, and cross-task generalizability. With this we intend to 1) providing an overview Dynamic CNN; 2) compare them across carious tasks of classification, segmentation and real time analysis; and 3) summarizing the key challenges and possible future research directions (see Fig. 1 for an overview).

These findings are useful contributions to the current research on CNN optimization, offering practical guidance on building efficient neural networks for diverse data modalities.

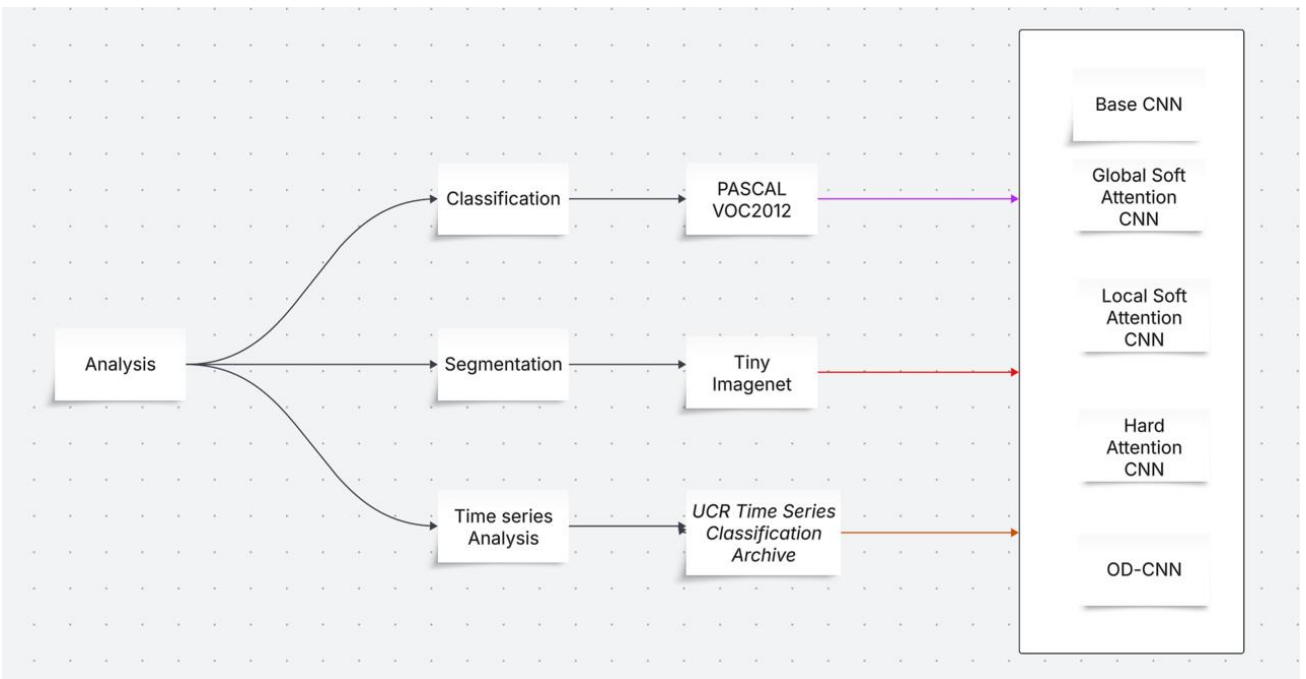

*Figure 1. Overview of our Analysis*

## 2. Methods

To fully evaluate the advancements and implementation of Dynamic CNNs, we utilized various dynamic CNN architectures across multiple datasets, systematically evaluating their accuracy and computational efficiency. The following sections provide an in-depth explanation of our methodology.

### 2.1. Model Architectures

Our project compares five CNN variants, each based on the ResNet18 backbone but differing in their approach to feature extraction and context adaptation using different attention. The models are:

#### 2.1.1. Base CNN

For our base model, we selected ResNet-18 due to its optimal balance between depth and computational efficiency. ResNet-18 is a variant of the ResNet family of architectures that incorporates the residual connections to combat the vanishing gradient issue while keeping a relatively low depth in comparison to classical implementations. This architecture requires significantly fewer computational resources compared to deeper models such as ResNet-50, thus making it particularly suitable for our experimental setup.

We Initially experimented using ResNet-50 proved to have infeasibly high computational costs, and this could potentially have restricted our ability to perform extensive comparisons across different model variants. ResNet-18 offered a good trade-off by delivering architectural consistency across all setups and also by maintaining computational demands within reasonable levels. In particular, every model was trained for 30 epochs, as discussed in the sections that follow, to ensure fair comparisons. The architecture of our ResNet-18 model is shown in Figure 2.

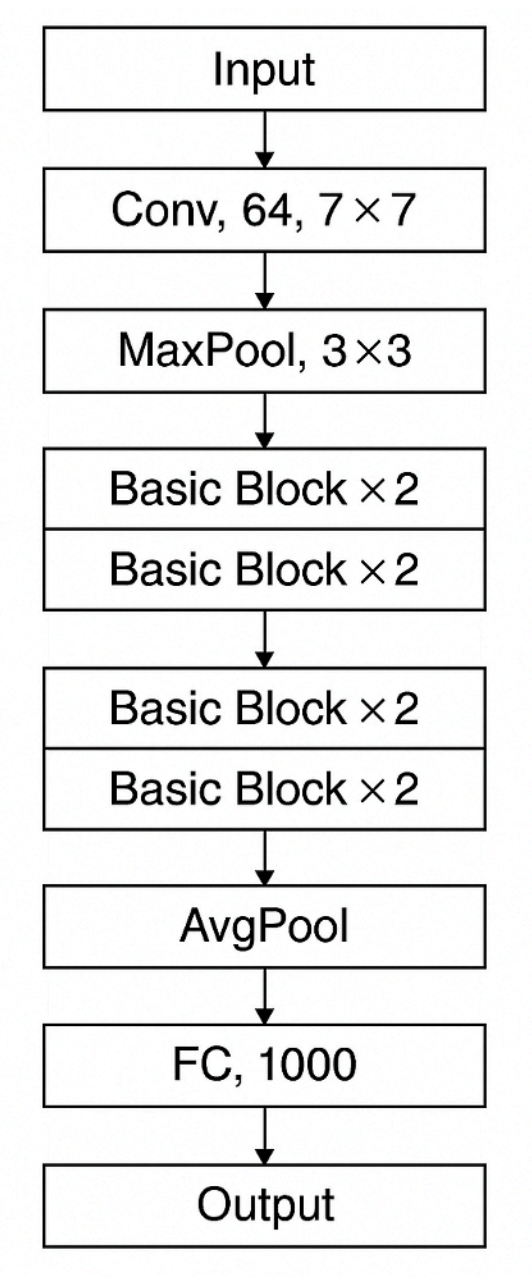

*Figure 2: ResNet-18 Architecture*

#### 2.1.2. Local Soft Attention CNN

To further enhance our dynamic CNN evaluation, we extended our analysis using Local Soft Attention, where attention is applied to per-pixel feature maps. This approach enables the network to dynamically focus on the most informative spatial regions within an image, providing substantial improvements in tasks such as semantic segmentation and rare event detection. Core mechanism of Local soft attention includes:

**Kernel Representation (KR):** A separate network maintains a dynamic representation of kernel decisions at each layer. This Kernel Representation is computed using three key components:

- Current input features are extracted using global average pooling, which captures the essential feature distribution.
- Historical Context**:** Kernel data from previous layers is integrated to maintain continuity across layers.
- Learnable Parameters: These are used for kernel generation, enabling the model to adapt kernel behavior dynamically.

This Kernel Representation is iteratively updated across layers, allowing it to learn and adapt based on the evolving feature distribution.

**Dynamic Kernel Generation:**
For each convolutional layer, a lightweight auxiliary network, known as the Kernel Generator, uses the Kernel Representation (KR) to compute attention weights (A).

These weights dynamically adjust the contribution of each parallel convolutional kernel in the layer. The final kernel for each layer is computed as a weighted sum:

$$W(Dynamic) = \Sigma AW \quad \text{(eq. 1)}$$

where, A are the attention weights determined via a Squeeze-and-Excitation mechanism, and W are the convolutional kernels.

This mechanism allows the model to selectively prioritize the most relevant features at each spatial location, enhancing its capacity for fine-grained attention.

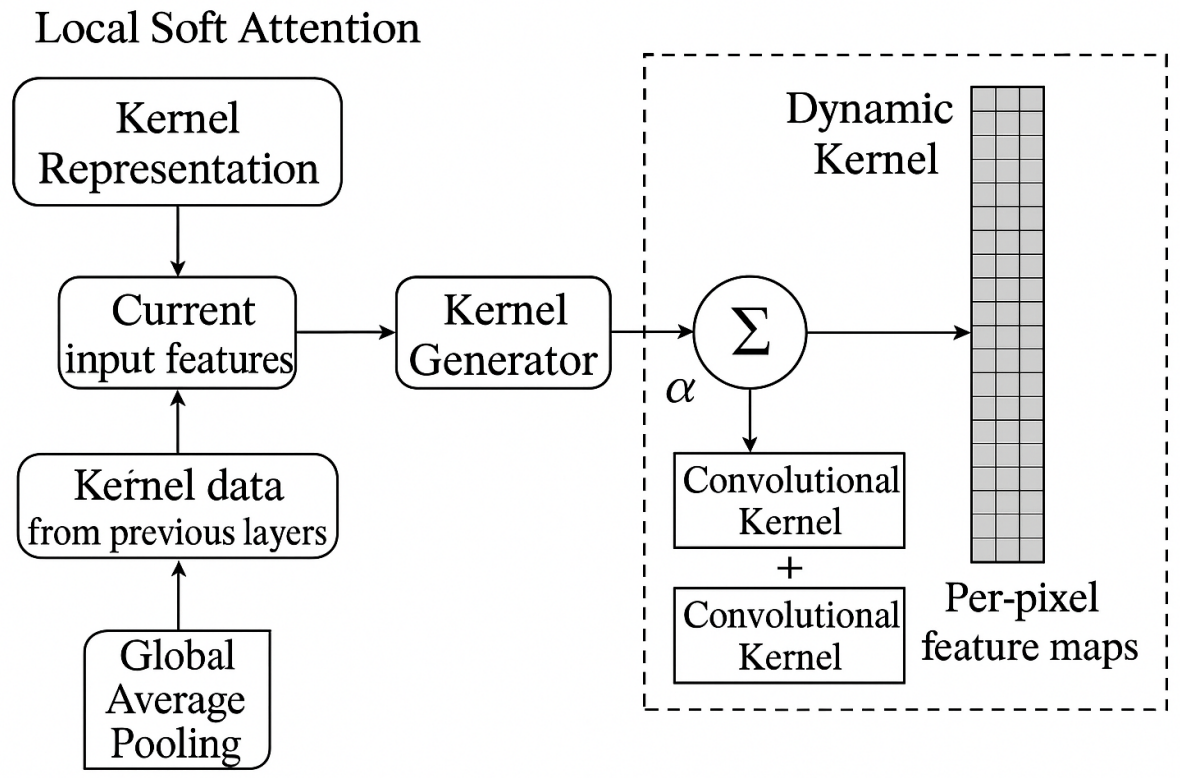

*Figure 3. Local Soft Attention Architecture*

2.1.3 Global Soft Attention CNN

The Global Soft Attention Convolutional Neural Network applies an attention mechanism across the full range of the input feature map to enable the network to adaptively emphasize the most salient features throughout the whole image. The approach is especially useful in applications where it is valuable to understand the general context of the image, for example, classification, object detection, and global context comprehension. Core mechanism of Global soft attention includes:

- **Full Feature Map Attention:** The model examines the complete feature map, utilizing all hidden states from the encoder to produce thorough attention weights.
- **Global Average Pooling:** The model uses global average pooling across all channels to create a context vector, capturing the overall feature distribution of the image.
- **Attention Weight Calculation:** A lightweight fully connected layer (or convolutional layer) generates attention scores for each channel. To which sigmoid is applied, ensuring they range between 0 and 1.
- **Feature Enhancement and Suppression:** The calculated attention weights are applied across the entire feature map. Wherein the important features are enhanced (multiplied by higher weights), while less important features are suppressed (multiplied by lower weights).

Mathematically we can use below equation to calculate Attention(A):

$$A = \text{sigmoid}(W.GAP(F)) \quad \text{(eq. 2)}$$

where, F: Feature map of image initially, W: The learnable weights and GAP: Global Average pooling

And the final feature map is calculated as:

$$F_new = A * F_old \quad \text{(eq. 3)}$$

where new Feature map (F_new) is the product of attention and old Feature map (F_old)

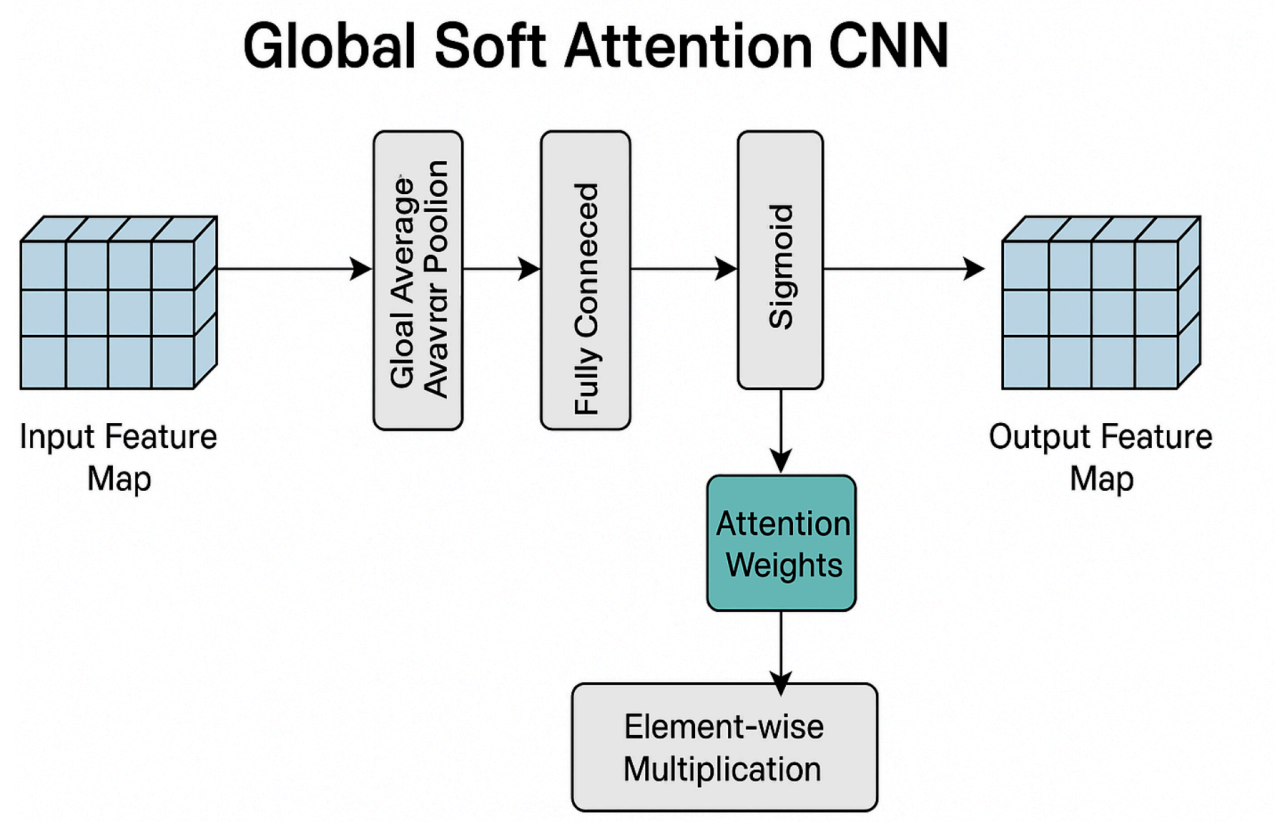

*Figure 4. Global Soft Attention Architecture*

2.1.4. Hard Attention CNN

For this model we extended the baseline ResNet18 convolutional neural network by adding a dynamic selection of convolutional kernels based on the particular input task, inspired by the MetaDOCK framework. We wanted to demonstrate improved performance and flexibility with minimal increases in computational costs. The working principles of the MetaDOCK framework can be summarize as:

- **Task-Specific Pruning**: MetaDOCK determines which kernels within a CNN are applicable for a provided input task. It prunes unnecessary kernels both during meta-training (across tasks) and task-specific adaptation (inner-loop updates).
- **Adaptive Kernel Selection**: Rather than directly applying gradients, MetaDOCK learns to dynamically switch kernels on or off to increase model flexibility for each task.
- **Two-Level Adaptation:** Adaptive Kernel Selection: Rather than directly applying gradients, MetaDOCK

learns to dynamically switch kernels on or off to increase model flexibility for each task.

- **Inner-Level**: Customizes kernels for each task at fine-tuning time, additionally optimizing the model according to input-dependent task features.

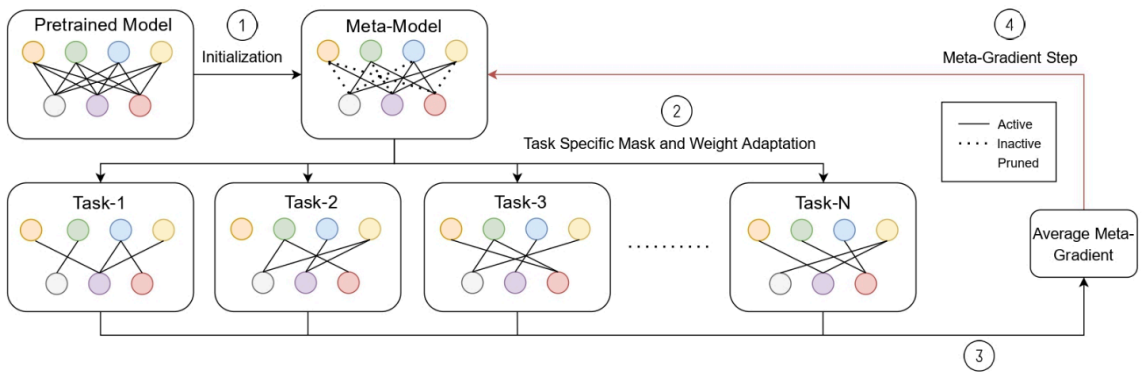

*Figure 5.* MetaDOCK *Hard Attention Architecture*

2.1.5. Omni Directional CNN

Omni-Directional Convolutional Neural Networks (ODCNNs) are a novel convolutional neural network architecture that addresses the inherent directional bias of conventional CNNs. As opposed to conventional CNNs, which generally utilize fixed-directional filters to scan data in pre-defined axes, i.e., horizontal and vertical ODCNNs incorporate new filters that are capable of extracting features at multiple orientations all at once. This is attained by a mechanism of rotated convolutional kernels or direction-adaptive units that enable the network to detect patterns irrespective of their orientation in the input data. The omni-directional structure significantly enhances the network's ability to identify rotation-invariant features, making ODCNNs highly effective for such applications where objects or patterns may appear at arbitrary angles, such as satellite image analysis, medical image processing, and object detection in unconstrained environments where conventional CNNs can fail due to their directional constraint.

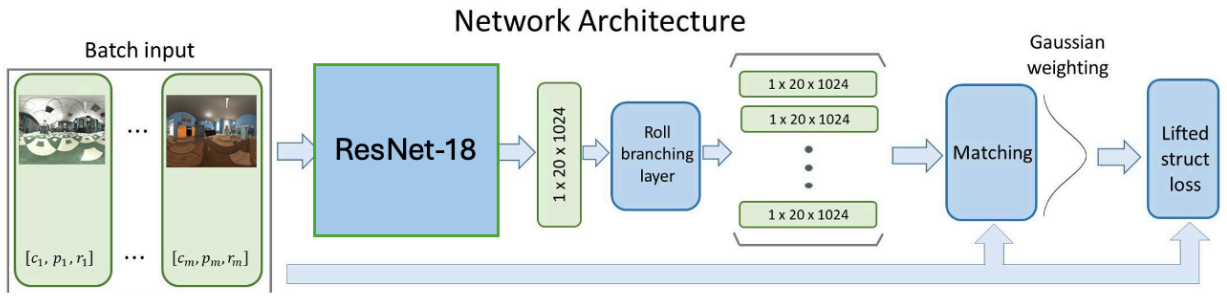

*Figure 6. Omni Directional CNN*

2.2. Datasets

We employed three datasets to test our model architectures on three different tasks. For image segmentation, we used the Pascal VOC 2012 dataset, an established benchmark containing 20 object categories and pixel-wise labeled images, to provide a good testing environment to assess semantic segmentation. For the classification of images, we used the Tiny ImageNet dataset, which is a miniature version of ImageNet containing 200 classes and images of size 64x64 pixels to enable efficient training and testing of classifiers. Finally, for time series analysis, we utilized the UCR Adiac dataset from the UCR Time Series Classification Archive, which contains 781 time series samples that belong to 37 various leaf shape classes. This makes it a good benchmark to test the time series analysis capability of our model. Collectively, these datasets allowed us to perform a comprehensive assessment of our model's efficacy in segmentation, classification, and time series-based tasks.

2.3. Training Strategy

To enable a fair and uniform evaluation across all versions of the model, we maintained standardized training settings with regard to the loss function, optimizer, batch size, and learning rate. Specifically, we used configuration as mentioned in Table 1.

| Model Parameters | Values |
|---|---|
| *Optimizer* | Adam |
| *Learning Rate* | 0.001 |
| *Batch Size* | 32 |
| *Epochs* | 30 |
| *Dropout* | 0.2 |
| *Loss Function* | categorical Cross-Entropy |

*Table 1. Model Configurations*

**Classification:** In the case of image classification experiments, we used ResNet18 as the base model, keeping the fundamental residual block design intact while making variant-specific changes for each model variant (Standard, Local Soft Attention, Global Soft Attention, and ODConv).

**Image Segmentation:** In our segmentation models were based on ResNet18. We employed Faster R-CNN as the segmentation framework due to its favorable trade-off between simplicity and performance. This allowed us to study the impact of various dynamic convolutional methods on segmentation accuracy without introducing additional complexity.

**Real Time Series Analysis:** For time series forecasting, we employed two custom models (Net1_DCNN and Net2_DCNN) designed to leverage Dynamic Convolutional Layers (DCNN). The layers were pre-initialized with pre-trained weights from a standard convolutional network so that the dynamic convolutional layers were pre-computed for speedy execution. Specifically:

- **Net1_DCNN** is made up of one dynamic convolutional layer followed by max pooling and fully connected layers, providing a light and efficient architecture.
- **Net2_DCNN** extends this architecture with one more convolutional layer, giving a deeper representation with parameter efficiency. The pre-trained dynamic convolutional layers were frozen after pre-training so that the models could concentrate on time-series specific feature extraction without unnecessary computational burdens.

This consistent training arrangement across classification, segmentation, and time series tasks enabled a robust and fair comparison of all model variants, highlighting the impact of dynamic convolutional approaches on performance.

2.4. Evaluation Metrics

To extensively evaluate the performance of our models across various tasks, we utilized a range of task-specific evaluation measures. To calculate computational performance, we approximated FLOPs (Floating Point Operations), which give a precise calculation of the computational cost of each model during inference. To evaluate the performance of image segmentation experiments on the Pascal VOC 2012 dataset, we employed Mean Intersection over Union (mIoU), a common metric that computes the average intersection over union of predicted segmentation masks and corresponding ground truth across all classes. To assess image classification performance on the Tiny ImageNet dataset, we evaluated Accuracy, thereby providing a stringent evaluation of model predictions. Furthermore, we used Multi-Fold Accuracy with Mean, with multiple runs of training and testing for time series classification on the UCR Adiac dataset. These measurements collectively provided us with an overall assessment of our model's computational speed, classification accuracy, and segmentation precision. The summary configuration is showcased in Table 2.

| Dataset | Operation | Samples | Classes |
|---|---|---|---|
| *Tiny ImageNet* | Classification | 100000 | 200 |
| *Pascal VOC 2012* | Segmentation | 11530 | 21 |
| *UCR Archive* | Time series analysis | Adiac | 781 |

*Table 2. Dataset Summary Table*

3. Experimental Results

In this section we will discuss our experiments and the results we obtained focusing on the performance of our attention-based models. We wanted to see how different attention mechanism (Local Soft Attention, Global Soft Attention, and Omni-Directional Convolution) performance change the accuracy and computational efficiency on different tasks (classification, segmentation, and time series). We first start with the classification tasks using Tiny ImageNet.

3.1. Classification

**Data Preprocessing:** We used the Tiny ImageNet dataset for image classification, which contains 200 classes, 500 training images, and 50 validation images, all of which were resized to 224x224 pixels. The data preprocessing pipeline utilized data augmentation via ImageDataGenerator, in order to facilitate a more robust training process:

```
train_generator = train_datagen.flow_from_directory(
    train_dir,
    target_size=(224, 224),
    batch_size=batch_size,
    class_mode='categorical'
)

val_generator = val_datagen.flow_from_directory(
    '/home/vikrant/DL_Project/tiny-imagenet-
200/tiny-imagenet-200/val/organized',
    target_size=(224, 224),
    batch_size=batch_size,
    class_mode='categorical'
)
```

*Code Sample: Data Pipelines for Validation and Train Data*

**Training Configuration:** On the classification models we used Cross-Entropy Loss, and we used the Adam Optimizer with a starting learning rate of 0.001. We also chose a batch size of 32 to balance speed of training and memory.

```
test_generator = test_datagen.flow_from_directory(
    '/home/vikrant/DL_Project/tiny-imagenet-
200/tiny-imagenet-200/val/organized',
    target_size=(224, 224),
    batch_size=32,
    class_mode='sparse',
    shuffle=False
)
```

*Code Sample: Data Pipelines for Test Data*

**Callback Mechanisms:** To ensure efficient training, we integrated a series of call back function:

```
callbacks = [

ModelCheckpoint('best_model_{epoch:02d}_{val_ac
curacy:.3f}.h5',
                save_best_only=True,
monitor='val_accuracy'),
    ReduceLROnPlateau(monitor='val_loss',
factor=0.1, patience=5, min_lr=1e-6),
    EarlyStopping(monitor='val_loss', patience=10)
]
```

*Code Sample: Callback Mechanism*

Below are the results for the accuracy outcome of all the model and the flop values.

| *Model* | *Accuracy* |
|---|---|
| *Base Model* | 65.2 |
| *Global Soft Attention* | 70.1 |
| *Local Soft Attention* | 71.5 |
| *Hard Attention* | 68.7 |
| *OD-CNN* | 73.4 |
| *Model* | FLOPs (In Billions) |
| *Base Model* | 1.5 |
| *Global Soft Attention* | 1.8 |
| *Local Soft Attention* | 2.0 |
| *Hard Attention* | 2.1 |
| *OD-CNN* | 2.3 |

*Table 3. Result for Classification Task*

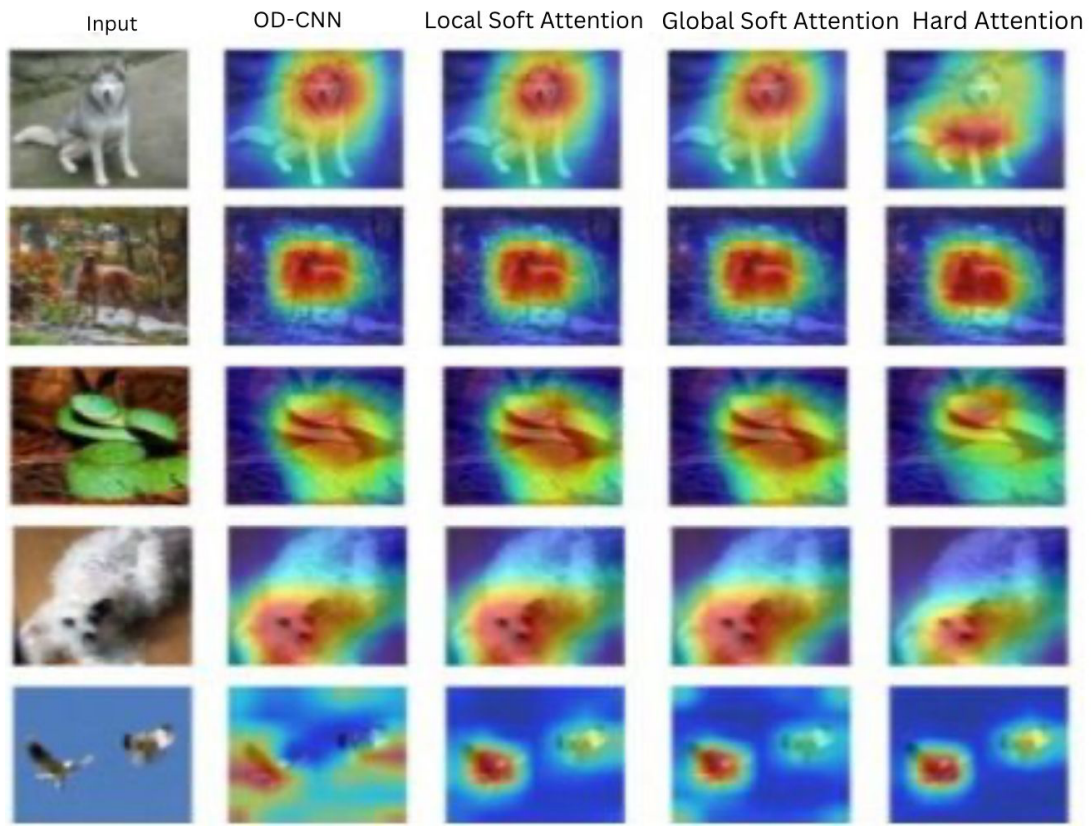

*Figure 7. Attention Maps from 4 models*

### 3.2. Segmentation

**Data Preprocessing:** We used the Pascal VOC 2012 dataset for the image segmentation, where images have annotations at the pixel level, 20 object categories from the dataset for this study. To improve the training process and model generalizability, the training data has undergone data augmentation using Keras ImageDataGenerator, which applies transformations including:

- Random horizontal flips.
- Random rotation (up to 15 degrees).
- Random scale and random zoom.

```
from tensorflow.keras.preprocessing.image import
ImageDataGenerator

train_datagen = ImageDataGenerator(
   horizontal_flip=True,
   rotation_range=15,
   zoom_range=0.2,
   brightness_range=[0.8, 1.2]
)
```

*Code Sample: Image Augmentation*

**Training Configuration:** For our segmentation we used the similar configuration as for classification with callback mechanism to save the best model. We used mIoU score to compare the models which classifies the images into 21 categories with 20 valid categories and 1 background.

$$mIoU = \frac{1}{c} \sum TP/(TP + FP + FN)$$

where, C is the number of classes, TP are True Positive case, FP is False Positive case and FN is false Negative

**Results:** Below are the final results for the mIoU results of all the model.

| *Model* | mIoU |
|---|---|
| *Base Model* | 67.5 |
| *Global Soft Attention* | 69.60 |
| *Local Soft Attention* | 70.17 |
| *Hard Attention* | 70.69 |
| *OD-CNN* | 73.09 |

*Table 4. mIoU for Segmentation Task*

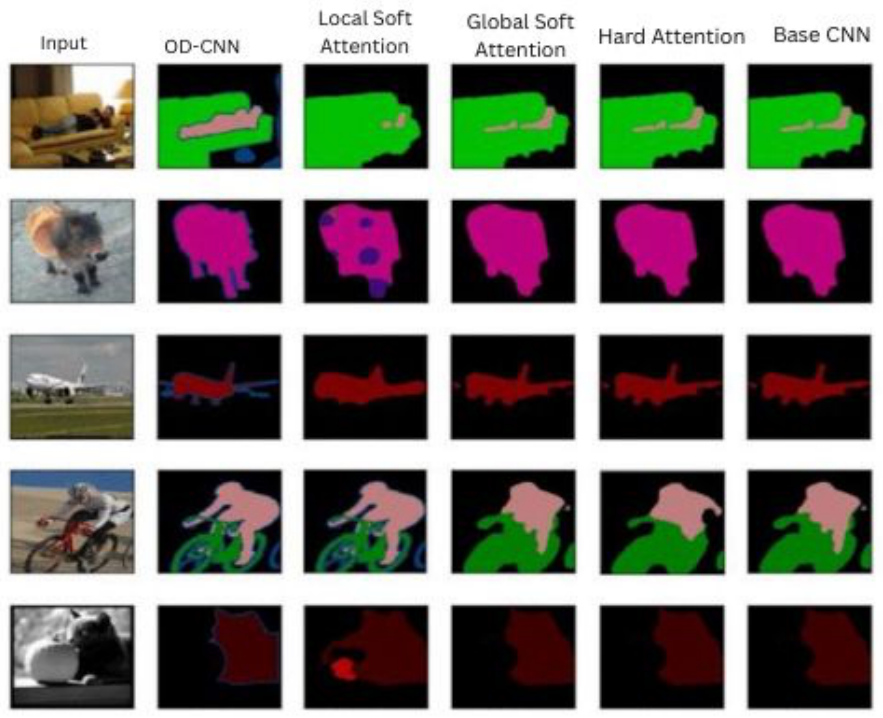

*Figure 8. Sample Segmentation output*

### 3.3. Time Series Analysis

For time series analysis we used Adiac from UCRArchive 2018 for time series analysis that consists of 781 time series samples. We trained our model on 10 folds and below were the results:

| *CNN (10-Fold Result)* | *Loss* | *Accuracy* |
|---|---|---|
| *0* | 1.461 | 0.506 |
| *1* | 1.447 | 0.564 |
| *2* | 1.475 | 0.526 |
| *3* | 1.472 | 0.462 |
| *4* | 1.398 | 0.654 |
| *5* | 1.526 | 0.590 |
| *6* | 1.398 | 0.590 |
| *7* | 1.457 | 0.551 |
| *8* | 1.560 | 0.615 |
| *9* | 1.429 | 0.654 |

| *D-CNN (10-Fold Result)* | Loss | Accuracy |
|---|---|---|
| *0* | 1.349 | 0.620 |
| *1* | 1.093 | 0.692 |
| *2* | 1.199 | 0.641 |
| *3* | 0.996 | 0.551 |
| *4* | 1.237 | 0.667 |
| *5* | 1.271 | 0.641 |
| *6* | 1.209 | 0.603 |
| *7* | 1.237 | 0.615 |
| *8* | 1.055 | 0.782 |
| *9* | 1.294 | 0.718 |

*Table 4a. Accuracy for Time Series Task*

| *Metric* | *CNN Model* | *D-CNN Model* |
|---|---|---|
| *Mean acc. CNN* | 0.571 | 0.653 |
| *Std. acc. CNN* | 0.059 | 0.062 |

*Table 4b. Statistic for Time Series Task*

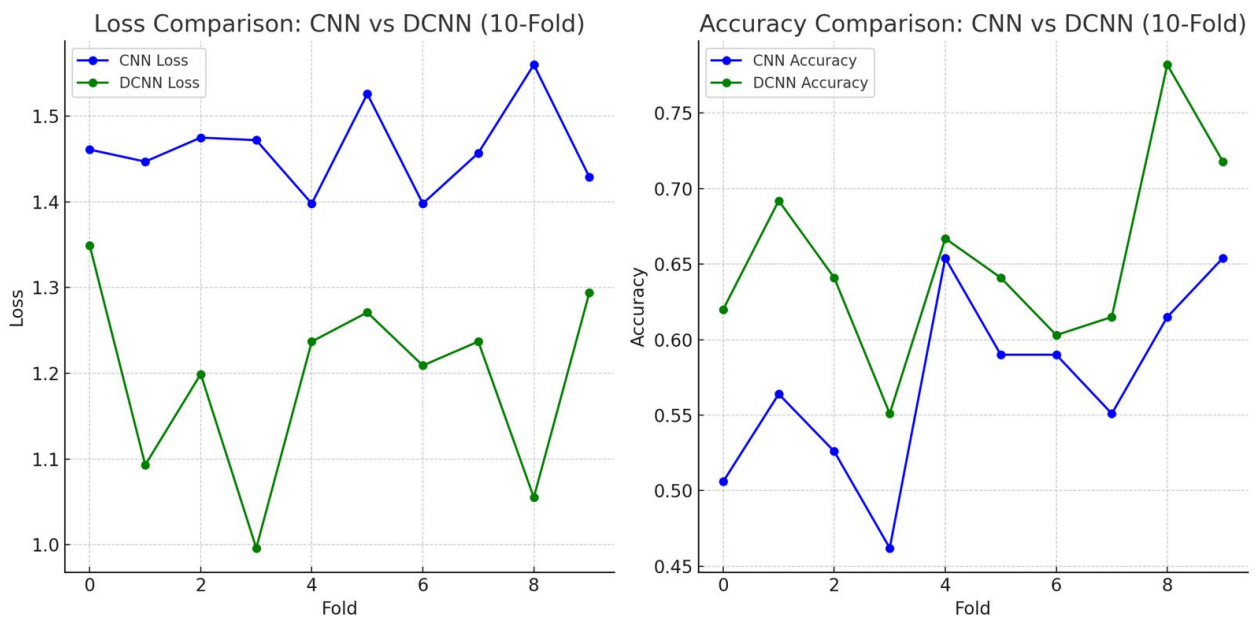

*Figure 9. Loss and Accuracy Comparison for CNN and DCNN*

## 4. Discussions

### 4.1. Performance Analysis by Task

**Classification Performance:** The attention-based model outperformed the base model and in attention-based model the Omni Direction CNN performed best. This is consistent with expectations because attention mechanisms supply models with more contextual information that allows them to perform better. However, I thought that local attention-based model will perform better because it has more contextual information and adjust the kernels based on the individual feature maps. But to my surprise the Omni direction CNN performed best. I think this was because the model the Omni Direction CNN is orientation-agnostic approach. As the final results Omni directional CNN was clear winner in terms of accuracy of 73.4% on Tiny ImageNet.

**Segmentation Task:** With the highest mIoU score of 73.09%, OD-CNN once again outperformed other models in segmentation on Pascal VOC 2012. Notable advancements were also made by the Local and Global Soft Attention models, highlighting the importance of adaptive focus on pertinent image regions.

**Time-Series Analysis:** With a higher accuracy (mean 0.653) than the Base CNN (mean 0.571), DCNN demonstrated a definite advantage in the time-series analysis task using UCR Adiac. The fact that this improvement was constant across all ten folds shows how effective dynamic convolution is at learning temporal patterns.

### 4.2. Impact of Attention Mechanisms

The results shows that both hard and soft attention mechanisms greatly improve model performance by enabling the network to concentrate on key areas in the input data. Hard Attention performed best in situations with distinct, clear object regions, while Local Soft Attention performed better in segmentation tasks than Global Soft Attention, probably because it could capture fine-grained details, and trained kernels were more specific than the soft attention kernels.

### 4.3. Computational Efficiency

Despite having the best performance, our analysis revealed that all the attention based require high computation cost. And specifically, OD-CNN had the highest computational cost (2.3 GFLOPs). But the notable performance improvements outweighed this trade-off. Despite having the highest computational efficiency, base CNN had the lowest accuracy on all tasks.

### 4.4. Summary of Results

When comparing results for accuracy, all the dynamic models were significantly better than the base model. This is consistent with expectations because attention mechanisms supply models with more contextual information that allows them to perform better at classification and segmentation tasks. The pattern is strongly evident in our findings.

Surprisingly, our Omni-Directional CNN outperformed all attention-based approaches. This better performance is due to OD-CNN's ability to process features of various orientations simultaneously, contrary to the traditional kernel-based attention that focuses attention on small regions only.

In terms of computational cost, although the baseline model took the least number of FLOPs (1.5B), the global attention model with 1.8B FLOPs presented the best tradeoff between computation and performance gain. Global feature map computations are more efficient because they operate on complete feature spaces rather than restricted areas.

In conclusion, our results depict a clear performance gain with increasing computational complexity. Whereas the baseline model saturated the performance limits, attention-based models showed consistent improvement, with OD-CNN achieving the best level of performance metrics at an acceptable computational cost (2.3B FLOPs).

### 4.5. Future Work

More dynamic model optimization for quicker inference may be investigated in future studies. A more thorough grasp of their potential might also be obtained by broadening the study to incorporate other dynamic architectures and more sophisticated attention mechanisms (like multi-head self-attention).

## 5. Contributions

This project was a major learning for me in terms of implementation and application of various CNN architectures. During the project, I made the following contributions:

**Data Preprocessing:** Initially, I started with preprocessing data that was used to train and test three major datasets used in this project: Tiny ImageNet for image classification, Pascal VOC 2012 concerning image segmentation, and the UCR Archive concerning real-time series analysis. Preprocessing was a crucial step to compare models fairly and improve performance results. As for Omni-Directional CNN testing, I augmented, resized, and rotated the test and training data to highlight the orientation-invariance property of the model. For Pascal VOC and tiny ImageNet datasets, I normalized segmentation masks and pixel-wise annotations into category-specific channels for more efficient training while loading the data with batches of 32.

**Contributions to Model Design and Training:** After the data preprocessing pipelines were in place, I developed and trained the Soft Attention-based models and Omni-Directional CNN (ODCNN) models. This involved incorporating the attention mechanisms, setting up the architectures, and optimizing the training procedure to deliver high performance on classification, segmentation, and time-series tasks.

For attention-based models, I implemented:

- Global Soft Attention: Channel-wise attention mechanism that rebalances feature importance across the entire feature map.
- Local Soft Attention: A spatial attention variant that attends to certain regions of interest, boosting performance by capturing more fine-grained contextual information.

In addition to the soft attention-based model, I developed the Omni-Directional CNN. Executed orientation pooling processes that combine features from a range of direction filters.

**Analyzing the Real Time series:** Once we were done with image classification and segmentation tasks, we also wanted to verify whether a dynamic CNN will perform better on time series data, as CNNs are mostly used deep learning architecture for time series analysis. For this, I built 2 network-based CNN architectures for which I downloaded data from the UCR archive for real-time analysis for end-to-end performance evaluation.

**Performance Evaluation:** Once all the models were trained, I performed performance assessment on the trained models by calculating and comparing values such as Mean Intersection over Union (mIoU) for segmentation, accuracy for classification, and Floating-Point Operations (FLOPs) for computation efficiency. My analytical contributions were:

- Benchmark Metric Implementation: The standardized evaluation metrics were applied to all model variants, allowing a comparison of the base model, attention-based approaches, and novel OD-CNN architecture on a level playing field.
- UCR time series data: I used a 10-fold performance evaluation to measure the robustness of the Dynamic CNN compared to the base CNN using statistical techniques like Standard Deviation and variance.

In conclusion, I worked in various capacities for the project, from data preparation to model training, and then model analysis. I mainly evaluated Omni-directional CNN, global soft attention, and local soft attention. Finally, I built and analyzed the time series-based model for real-time analysis.